\documentclass[conference]{IEEEtran}
\usepackage{graphics} 
\usepackage{epsfig} 
\usepackage{mathptmx} 
\usepackage{times} 
\usepackage{amsmath} 
\usepackage{amssymb}  
\usepackage{color}  
\usepackage{xcolor}
\usepackage[numbers]{natbib}
\usepackage{multicol}
\usepackage[hidelinks, bookmarks=true]{hyperref}                    
\usepackage{etoolbox}
\usepackage[font=small,labelfont=bf]{caption}

\DeclareMathAlphabet{\mathpzc}{OT1}{pzc}{m}{it}

\usepackage{subcaption}

\makeatletter
\patchcmd{\@makecaption}
  {\scshape}
  {}
  {}
  {}
\makeatother
\definecolor{agreen}{rgb}{0.25, 0.51, 0.20}
\definecolor{john_blue}{rgb}{ 0.0627    0.5569    0.9098}

\begin{document}

\title{Blind Bipedal Stair Traversal \\ via Sim-to-Real Reinforcement Learning}
 \author{\authorblockN{Jonah Siekmann$^{*\dag}$, Kevin Green$^*$, John Warila$^*$, Alan Fern$^*$, Jonathan Hurst$^{*\dag}$}
 \authorblockA{
     *CoRIS Institute, Oregon State University\\
     \emph{\{greenkev, warilaj, afern, jhurst\}@oregonstate.edu}}
     $\dag$Agility Robotics\\
     \emph{\{jonah.siekmann\}@agilityrobotics.com}\\
 }


%

\maketitle

\begin{abstract}
Accurate and precise terrain estimation is a difficult problem for robot locomotion in real-world environments. Thus, it is useful to have systems that do not depend on accurate estimation to the point of fragility. In this paper, we explore the limits of such an approach by investigating the problem of traversing stair-like terrain without any external perception or terrain models on a bipedal robot. For such blind bipedal platforms, the problem appears difficult (even for humans) due to the surprise elevation changes. Our main contribution is to show that sim-to-real reinforcement learning (RL) can achieve robust locomotion over stair-like terrain on the bipedal robot Cassie using only proprioceptive feedback. Importantly, this only requires modifying an existing flat-terrain training RL framework to include stair-like terrain randomization, without any changes in reward function. To our knowledge, this is the first controller for a bipedal, human-scale robot capable of reliably traversing a variety of real-world stairs and other stair-like disturbances using only proprioception.

\end{abstract}

\IEEEpeerreviewmaketitle

\section{Introduction}
In order to be useful in the real world, bipedal and humanoid robots need to be able to climb and descend stairs and stair-like terrain, such as raised platforms or sudden vertical drops, which are common features of human-centric environments. 
The ability to robustly navigate these environments is crucial to getting robots to work with and alongside humans safely. Achieving this level of robustness on a bipedal platform is no easy task; while other platforms such as quadrupedal robots benefit from inherent stability due to multiple points of contact with the ground at a given time and the ability to stop and stand like a table, bipedal robots such as Cassie rely entirely on dynamic stability (essentially always existing in a state of falling). On stair-like environments, this is especially apparent due to the difficulty of recovery from missteps with only two legs. 

By contrast, robots with quadrupedal morphologies have been able to use proprioception alone to negotiate stairs \cite{lee2020learning, bledt2018cheetah}, and hexapedal robots have even been able to use open-loop control to ascend and descend stairs \cite{moore2001stable}. While planar bipedal robots have been shown to be able to reject disturbances like large unexpected dropsteps \cite{park2012finite}, the vast majority of approaches seeking to enable such robots to negotiate stairs in the real world require either accurate vision systems \cite{gutmann2004stair, albert2001detection, michel2007gpu} or operation in a carefully controlled laboratory environment \cite{caron2019stair, agility2019video, figliolini2001climbing}, meaning the robot is localized through a known start location or the stairs are designed in tandem with robot morphology.

\begin{figure}[t!]
\centering
\includegraphics[width=0.45\textwidth]{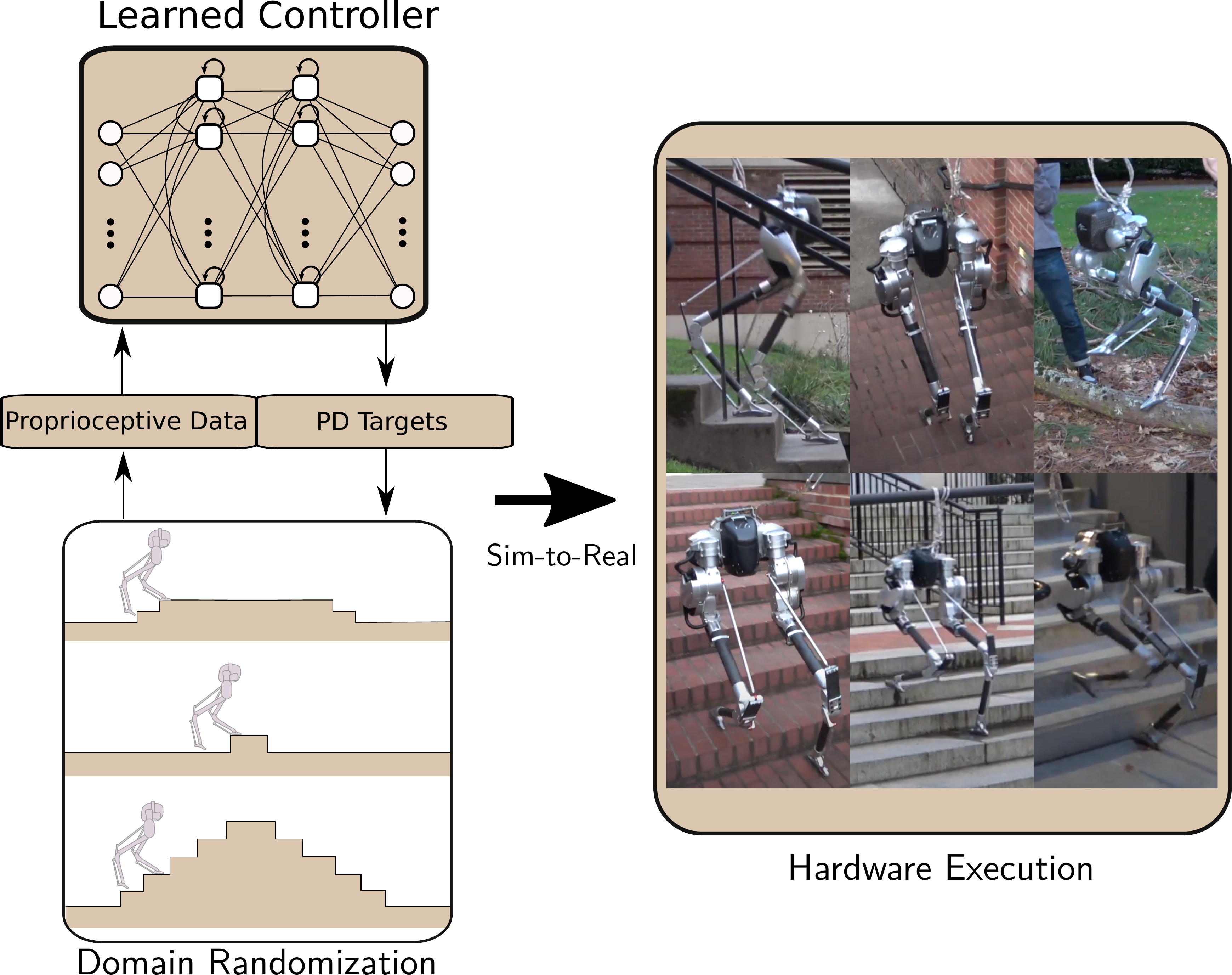}
\caption{In this work, we investigate the limits of blind bipedal locomotion. We present a training pipeline which produces policies capable of blindly ascending and descending stairs in the real world. These policies learn proprioceptive reflexes to reject significant disturbances in ground height, resulting in highly robust behavior to many real-world environments.}
\label{fig:title_fig}
\end{figure}

However, robots must be able to operate outside of controlled laboratory conditions and handle the massive variety of conditions in the real world. This goal is not compatible with a complete reliance on exteroceptive sensors such as RGB and depth cameras for accurate terrain estimation, which introduce fragility to real world conditions \cite{Focchi2020variablesensor}. For instance, cameras may be unreliable if exposed to occlusion, fog, or varying lighting conditions. Further, integrating a state-of-the-art computer vision system into a high-speed controller is technically difficult, especially on a computationally limited platform like a mobile robot. For practical purposes, underlying controllers should be as robust as possible while relying on as little information about the world as possible. Ideally, a bipedal robot should be able to traverse as much of the entire breadth of human environments as possible using proprioception, while relying on exteroceptive sensing for further efficiency and high-level planning (and being robust to mistaken perception). This begs the question: how robust can a blind bipedal robot be?


Reinforcement learning (RL) based approaches have begun to show significant promise at robust real-world legged locomotion \cite{lee2020learning, xie2018feedback, siekmann2020simtoreal}. Unlike optimization or heuristic-based control methods which rely on prescribed ground contact schedules or force-based event detection, RL can produce control policies which learn proprioceptive reflexes and strategies for dealing with unexpectedly early or late contact and rough terrain through exposure to a variety of disturbances during training. 
However, the limits of this approach are unclear and prior work has not been demonstrated on the scale and variety of disturbances involved in stair-like terrain. 

In this work, we show that robust proprioceptive bipedal control for complex stair-like terrain can be learned via an existing RL framework with surprisingly little modification. In particular, the only adjustment needed is the terrain randomization used during training, where we define a distribution over upward and downward going stairs including variation in height, width, and slope of the contact planes. Learning on this distribution allows for blind locomotion up and down unknown stairs as well as handling more general stair-like terrain characteristics, e.g. logs, curbs, dropoffs, etc. The learned controller is demonstrated in simulation and a variety of real-world settings. To our knowledge, this is the first demonstration of its kind and suggests the continued exploration into the limits of robust proprioceptive bipedal control. 

\section{Reinforcement Learning Formulation}
\label{sec:method}

We follow a sim-to-real reinforcement learning (RL) approach for learning bipedal locomotion and assume basic familiarity with RL \cite{Sutton2018}. In the general RL setting, at each discrete time step $t$ the robot control policy $\pi$ receives the current state $s_t$ and returns an action $a_t$, which is applied and results in a transition to the next state $s_{t+1}$. The state transition dynamics are unknown to the robot and are governed by a combination of environmental conditions, such as terrain type, and the robot dynamics. In addition, during learning, each state transition is associated with a real-valued reward $r_t$. The reward is governed by the application goals to encourage the desired behavior during learning. The RL optimization objective considered in this work is to learn a policy through interaction with the environment that maximizes the expected cumulative discounted reward over a finite-horizon $T$. That is, find a policy $\pi$ that maximizes: $J(\pi)=\mathbb{E}\left[\sum_{t=0}^{T}\gamma^t R_t\right]$, where $\gamma \in [0,1]$ is the discount factor and $R_t$ is a random variable representing the reward at time $t$ when following $\pi$ from a state drawn from an initial state distribution. 

For complex environments, RL typically requires large amounts of training experience to identify a good policy. Further, for biped locomotion, the training will involve many falls and crashes, especially early in training. Thus, training from scratch in the real-world is not practical and we instead follow a sim-to-real RL paradigm. Training is done completely in a simulation environment, with dynamics randomization (see below), and the resulting policy is then used in the real-world. 

In the remainder of this section, we detail the specific sim-to-real RL formulation used in this work, which follows recent work \cite{siekmann2020simtoreal} on learning different biped gaits over flat terrain. Surprisingly, only minimal changes were required to enable policy learning for the much more complex stair-like terrains of this paper\footnote{This was only discovered after a careful ablation analysis of our first success on stair-like terrain, which originally included seemingly necessary modifications to prior work, such as more complex reward functions and state features.} In particular, the only major modification required was the randomized domain generation of stair-like rather than mostly flat terrain as discussed later in Section \ref{sec:domain_rand}; no novel stair-specific reward terms were needed.

\subsection{State Space}
\label{subsec:statespace}

The state $s_t$ that is input to the control policy at each time step includes three main components. First, the state contains information about the robot's instantaneous physical state, including the pelvis orientation in quaternion format, the angular velocity of the pelvis, the joint positions, and the joint velocities. The second component of $s_t$ is composed of command inputs, which come from a human operator. These commands are subject to randomization during training to give the policies a wide breadth of experience attempting to traverse stairs over a variety of speeds and approach angles. Details of this randomization can be seen in Table \ref{table:command_rand}.

\begin{table}[!h]
\centering
\begin{tabular}{|lll|}
\hline
Command & Probability of Change & Range \\ \hline
Forward Speed &  1/300 & [-0.3m/s, 1.5m/s] \\
Sideways Speed & 1/300 & [-0.3m/s, 0.3m/s \\ 
Turn Rate & 1/300 & [-90deg/s, 90deg/s] \\ \hline
\end{tabular}
\caption{At each timestep, each command input to the policy is subject to a 1/300 probability of being altered. When this occurs, a new command is sampled from a uniform distribution parameterized by the rightmost column. Given that maximum episode length is 300 discrete timesteps, this means each command will change at least once on average per episode.}
\label{table:command_rand}
\end{table}

The third component includes two cyclic clock inputs, each corresponding to a leg of the robot, $p$:
\begin{equation}
 p =  \left\{
\begin{array}{ll}
    \sin\left(2 \pi (\phi_t + 0.0)\right) \\
    \sin\left(2 \pi (\phi_t + 0.5)\right) \\
\end{array} 
\right.
\end{equation}
Here $\phi_t$ is a phase variable which increments from 0 to 1, then rolls back over to 0, keeping track of the current phase of the gait. The constant offsets $0.0$ and $0.5$ are phase offsets used to make sure that the left and right legs are always diametrically opposite of each other in terms of phase during locomotion.




\subsection{Action Space}
\label{subsec:actionspace}

The output action $a_t$ of the control policy at each time step (running at 40Hz) is an 11 dimensional vector with the first 10 entries corresponding to PD targets for the joints, each of which are fed into a PD controller for each joint operating at 2KHz. Prior work has found it advantageous to learn actions in the PD target space rather than directly learning the higher-rate actuation commands \cite{peng2016learning}.

The final dimension of $a_t$ is a clock delta $\delta_t$ (refer to \ref{subsec:statespace} for information on clocks), which allows the policy to regulate the stepping frequency of the gait. Intuitively, this allows the controller to choose an appropriate stepping frequency for a particular gait, command, and terrain. Specifically, the phase variable $\phi$ in the state representation (Section \ref{subsec:statespace}) is updated at each timestep $t$ by, 
\begin{equation}
\phi_{t+1} = \text{fmod}(\phi_{t} + \delta_t, 1.0).
\end{equation}
This delta is bounded in a way such that the policy can choose to regulate the gait cycle between 0.5x and 1.5x the nominal stepping frequency (which is approximately one gait cycle every 0.7 seconds). While this component is included in the control policy action, it does not appear to have a large impact on performance and the learned policy does not vary $\delta_t$ much in response to disturbances. We suspect that future ablation analysis will show that it is not important for performance on the real robot.\footnote{We leave this as a hypothesis here, since we have not been able to test on the real robot at the time of submission.}

\subsection{Reward Function}
We use the method introduced in \cite{siekmann2020simtoreal} to specify our reward function. To briefly review this method, we desire a reward framework which allows for penalizing the policy for large magnitudes of some quantities of the environment at certain times, while permitting those quantities to be large at other times. We designate foot forces and foot velocities as two such quantities; punishing foot forces incentivizes the policy to lift the foot, while punishing foot velocities incentivizes the policy to place the foot. We add additional cost terms on top of these foundational reward terms, including a cost incentivizing the policy to match a translational velocity and orientation. We also employ costs which encourage smooth actions, energy efficiency, and to reduce pelvis shakiness. For a detailed explanation of the reward function used, see the Appendix. As in \cite{siekmann2020simtoreal}, we do not rely on expert reference trajectories to learn behaviors.

\subsection{Dynamics Randomization}
\label{sec:dynamicsrand}
In order to overcome any modeling errors that may be present in our simulated Cassie environment, we randomize several important quantities of the dynamics at the beginning of each episode during training as in previous work \cite{peng2019simtoreal} \cite{siekmann2020simtoreal}. These randomized parameters are listed in Table \ref{table:dynamicsrand}.

\begin{table}[!h]
\centering
\begin{tabular}{|l|l|l|}
\hline
Parameter & Unit & Range \\ \hline
 Joint damping                      & Nms/rad   &   $[0.5, 3.5] \times \text{default values}$ \\ \hline
 Joint mass                         & kg        &   $[0.5, 1.7] \times \text{default values}$     \\ \hline
 Ground Friction                    & --   &   $[0.5, 1.1] $     \\ \hline
 Joint Encoder Offset               & rad  &   $[-0.05, 0.05]$     \\ \hline
 Execution Rate               & Hz  &   $[37, 42]$     \\ 

 \hline

\end{tabular}
\caption{To prevent overfitting to simulation dynamics and facilitate a smooth sim-to-real transfer, we employ dynamics randomization. The above ranges parameterize a uniform distribution for each listed parameters. Damping, mass, friction, and encoder offset are randomized at the beginning of each rollout, while execution rate is randomized at each timestep to mimic the effect of variable system delay on the real robot.}
\label{table:dynamicsrand}
\end{table}

\subsection{Policy Representation and Learning}

We represent the control policy as an LSTM recurrent neural network \cite{hochreiter1997long}, with two recurrent hidden layers of dimension 128 each. We opt to use a memory-enabled network because of previous work demonstrating a higher degree of proficiency in handling partially observable environments \cite{heess2015memorybased} \cite{peng2019simtoreal} \cite{siekmann2020learning}. For ablation experiments, involving non-memory-based control policies, we use a standard feedforward neural network with two layers of dimension 300, with tanh activation functions, such that the number of parameters is approximately equal to that of the LSTM network. 

For sim-to-real training of the policy, we use Proximal Policy Optimization (PPO) \cite{schulman2017proximal}, a model-free deep RL algorithm. Specifically, we use a KL-threshold-termination variant, wherein each time the policy is updated, the KL divergence between the updated policy and the former policy is calculated and the update is aborted if the divergence is too large. During training, we make use of a mirror loss term \cite{mig2019symmetry} in order to ensure that the control policy does not learn asymmetric gaits. For recurrent policies, we sample batches of episodes from a replay buffer as in \cite{siekmann2020learning}, while for feedforward policies we sample batches of timesteps. Each episode is limited to be 300 timesteps, which corresponds to about 7.5 seconds of simulation time.




\begin{figure}[!t]
\centering
\includegraphics[width=1.0\columnwidth]{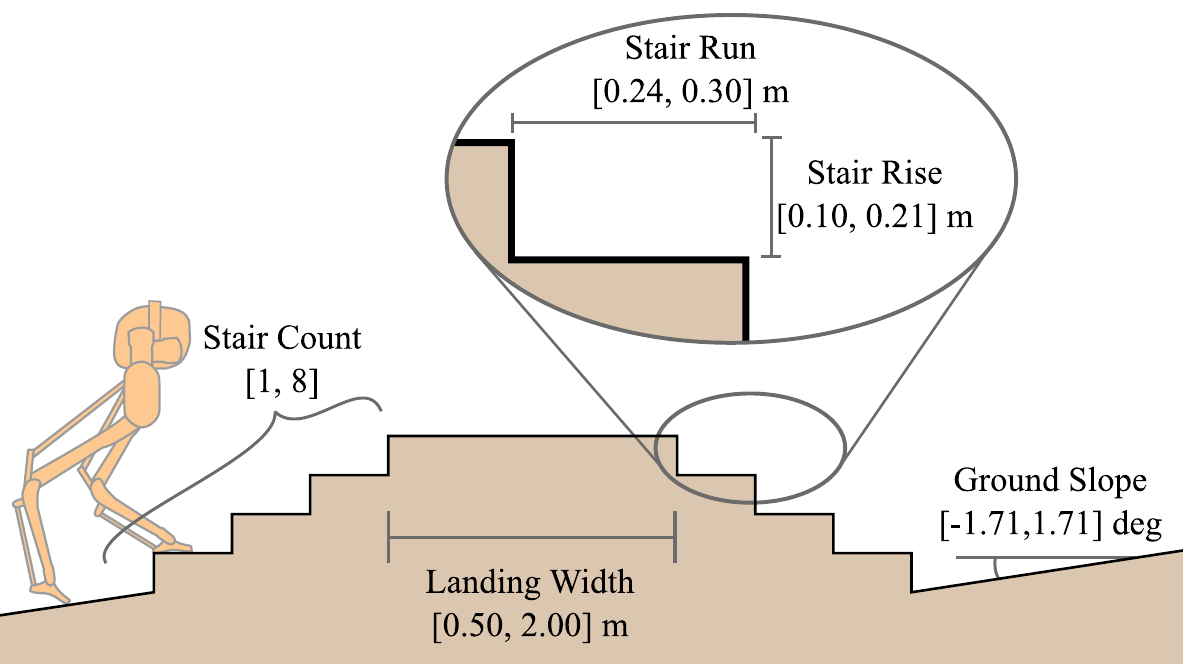}
\caption{In order to ensure robustness over a variety of possible stair-like terrain, we randomize a number of parameters when generating stairs at the start of each episode in simulation. These parameters include the number of stairs, the height of each stair, the length of each stair, the length of the landing atop the stairs, and the slope of the ground immediately before and after the stairs.}
\label{fig:stair_diagram}
\end{figure}

\section{Terrain Randomization}
\label{sec:domain_rand}

Previous work on applying RL to Cassie has either trained on flat ground \cite{xie2018feedback} \cite{siekmann2020learning} or on randomized slight inclines \cite{siekmann2020simtoreal}. Other work in applying deep RL has investigated employing a curriculum of rough terrains which become increasingly difficult as training progresses \cite{lee2020learning}. For the purpose of simplicity, we find that training on interactions with a randomized staircase without a curriculum is sufficient to learn robust behavior.

To this end, we train on a plane whose incline is randomized at the beginning of each rollout in the pitch and roll axes. This incline is between -0.03 radians and 0.03 radians. As part of the dynamics randomization, ground friction is randomized, increasing the potential difficulty of the environment.
The starting position of the stairs are randomized at the beginning of each rollout, such that the episode can start with the policy already on top of the stairs, or with the stairs up to 10 meters in front of the policy. 
This is done in order to ensure that the policy is able to see lots of experience on flat or inclined ground, as well as on stairs.

The dimensions of the stairs are randomized within typical city code dimensions, with a per-step rise of between 10cm and 21cm, and a run of 24cm to 30cm. The number of stairs is also randomized, such that each set of stairs has between 1 and 8 individual steps. A small amount of noise ($\pm$ 1cm) is added to the rise and run of each step such that the stairs are never entirely uniform, to prevent the policy from deducing the precise dimensions of the stairs via proprioception and subsequently overfitting to perfectly uniform stairs.

\section{Results} 
\label{sec:results}

We trained four groups of policies, each containing five policies initialized with different random seeds. First, we trained a group of simple LSTM policies with stair terrain randomization; these are referred to in this section as \textbf{Stair LSTM}. To investigate the importance of memory, we trained a group of feedforward policies also with stair terrain randomization; we denote these \textbf{Stair FF}. We also trained a group of policies without stair terrain randomization, and denote these \textbf{Flat Ground LSTM}, to investigate the importance of the terrain randomization introduced in this work.  The final group was trained with a simple additional binary input informing the policy whether or not stairs were present within one meter of the policy, referred to here as \textbf{Proximity LSTM}, in order to investigate the benefit of leaking information about the world to the policies.

\begin{figure}
\centering
\includegraphics[width=0.7\columnwidth]{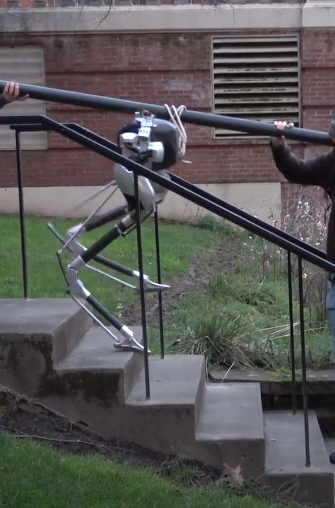}
\caption{The learned policies exhibit a high degree of blind robustness to a variety of stair-like terrain, and can reliably ascend and descend stairs of typical dimensions found in human environments.}
\label{fig:real_stairs}
\end{figure}

Each policy was trained until 300 million timesteps were sampled from the virtual environment, simulated with MuJoCo \cite{todorov2012mujoco}. Our selection of hyperparameters for the PPO algorithm includes a replay buffer size of 50,000 timesteps, a batch size of 64 trajectories for recurrent policies, and a batch size of 1024 timesteps for feedforward policies. Each replay buffer is sampled for up to five epochs, with optimization terminating early if the KL divergence reached the maximum allowed threshold of 0.02. We clear our replay buffer at the start of each iteration. We use the Adam \cite{kingma2014adam} optimizer with a learning rate of $0.0005$ for both the actor and critic, which are learned separately and do not share parameters.

\subsection{Simulation}



\noindent
\subsubsection{Probability of Successfully Ascending and Descending Stairs}

To understand the importance of memory and terrain randomization, we evaluate three groups of policies on the task of successfully climbing and descending a set of stairs in simulation. We compare the performance of Stair FF, Stair LSTM, and Flat Ground LSTM policies on this task. 

Specifically, we run 150 trials testing how often a policy is successfully able to climb a set of stairs with five steps, each with a tread of 17cm and a depth of 30cm (a typical real-world and relatively mild stair geometry).
This should give us an estimate of how reliably each group of control policies can climb a flight of stairs that it approaches blindly.
Success is defined as reaching the top of the flight of stairs without falling. We also apply this procedure for descending stairs, running 150 trials on stairs with the same dimensions, and record the rate at which each group of policies can reach the bottom without falling.

\begin{figure}
\centering
\includegraphics[width=0.95\columnwidth]{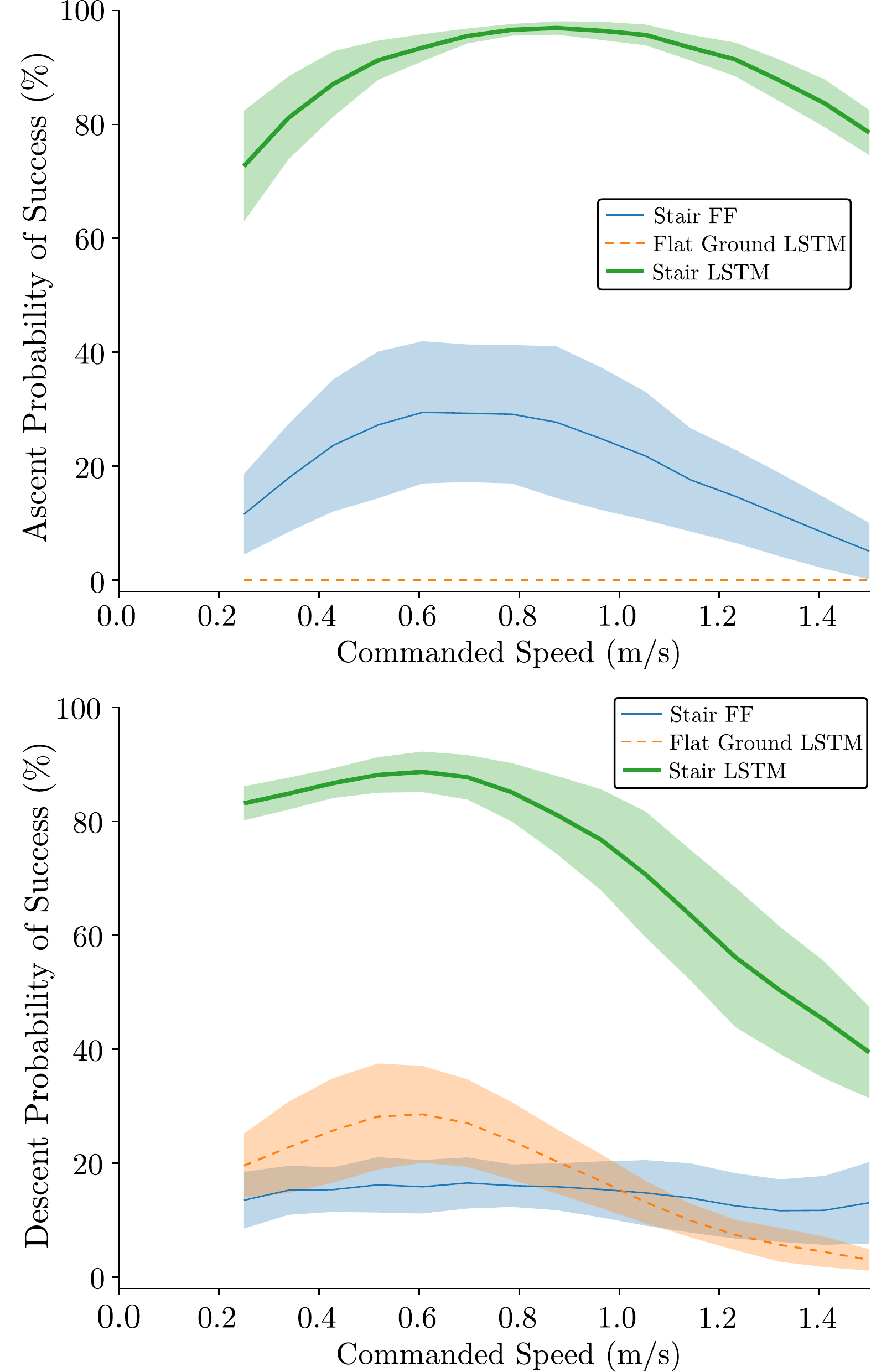}
\caption{We evaluate the probability of successfully climbing and descending stairs without falling as a function of commanded speed between 0.25 m/s and 1.5 m/s over 150 trials. For Stair LSTM policies, there seems to be an optimal approach speed for climbing stairs and a separate optimal descent speed. Stair FF policies do not attain high performance, implying that memory could be an important component of the learned behavior. Flat Ground LSTM policies, having never encountered stairs in training, are virtually unable to climb stairs while finding some success in descending stairs without falling over.}
\label{fig:sim_success}
\end{figure}

The results of these tests for three different training conditions is shown in figure \ref{fig:sim_success}. We note that the Stair LSTM policy has the highest overall probability of success. Nevertheless, the probability of success is dependent, in large part, on approach speed. The policies experience higher rates of failure at low speeds, where they may lack the momentum to propel themselves past poorly chosen foot placements. They also experience higher rates of failure at high speeds, possibly due to the more dynamic nature of high-speed gaits.

The Flat Ground LSTM policies, having never seen stair-like terrain during training, are unable to compensate and experiences a high rate of failure for both ascent and descent. The Stair FF policies, despite encountering stairs during training, are unable to learn an effective strategy for handling stairs, implying that memory may be an important mechanism for robustness to stair-like terrain.

\noindent
\subsubsection{Energy Efficiency Comparison}

To understand the consequences of training with terrain randomization, we also compare the cost of transport between Flat Ground LSTM policies, Stair LSTM policies, and Proximity LSTM policies. 
The cost of transport (CoT) is a common measure of efficiency of legged robots, humans and animals.
It is the energy used per distance, normalized by weight to be unitless. 
It is defined as 
\begin{equation}
    \text{CoT} = \frac{E_{\text{m}}}{Mgd},
\end{equation}
where $E_{\text{m}}$ is the energy used by the motors, $M$ is the total mass of the robot, $g$ is the gravitational acceleration and $d$ is the distance traveled.
The energy used by Cassie is calculated using positive actuator work and resistive losses via
\begin{equation}
    E_{\text{m}} = \int_{0}^{T} \Big(\sum_i max(\tau_i \cdot \omega_i, \; 0) + \frac{\omega_i^\text{max}}{P_i^\text{max}}\tau_i ^2 \Big)dt.
\end{equation}
Here $\tau_i$ is the torque applied to motor $i$ and $\omega_i$ is its rotational velocity.
We use two parameters to define the resistive losses in terms of torque, $P_i^{max}$ is the maximum input power and $\omega_i^{max}$ is the maximum speed of motor $i$.
The results of testing steady state CoT at 1 m/s on flat ground can be seen in Table \ref{table:cot}.
These calculations of CoT do not include the overhead power draw from computation and control electronics so they should not be used to compare between robots, only between control policies.

\begin{table}[!h]
\centering
\begin{tabular}{|lll|}
\hline
Policy Group & Mean CoT & Std. CoT \\ \hline
Proximity LSTM (stairs)&  0.47 & 0.0086 \\
Stair LSTM & 0.46 & 0.0323 \\ 
Proximity LSTM (flat) & 0.39 & 0.0257 \\
Flat Ground LSTM & 0.38 & 0.0205 \\ \hline
\end{tabular}
\caption{Locomotion efficiency as measured by cost of transport (CoT) for walking at 1 m/s over flat ground in simulation between three groups of policies over all five random seeds. We note that policies not trained on stair terrain randomization tend to learn more energy efficient gaits, though some energy efficiency can be recovered by providing the stair-trained policies with a binary stair presence/absence input.}
\label{table:cot}
\end{table}

We find that Flat Ground LSTM policies learn the most energy efficient gaits for walking on flat ground. Stair LSTM policies learn less efficient flat-ground gaits in order to be robust to stairs; however, the stair proximity input to the Proximity LSTM can help to recover some of this lost energy efficiency by allowing the learned controller to switch between a stair-ready gait and a more energy efficient, flat-ground gait. 
\subsection{Behavior Analysis}

To understand the strategy adopted by the policy, we can benefit from taking the perspective of experimental biology.
We specifically look at the behavior as the robot contacts the first step up or down after walking along flat ground.
First we will analyze the swing leg motion to understand how the robot places its foot on step ups and step downs.
Once the swing foot contacts a step up or down, the force applied by the foot on the ground during stance phase can be modulated to better prepare the robot for future steps.
We analyze how the ground reaction force and total impulse varies in the case of step ups and step downs.

\subsubsection{Swing Foot Motion}
\begin{figure}
\centering

\begin{subfigure}[b]{1.0\columnwidth}
\centering
\includegraphics[width=0.98\columnwidth]{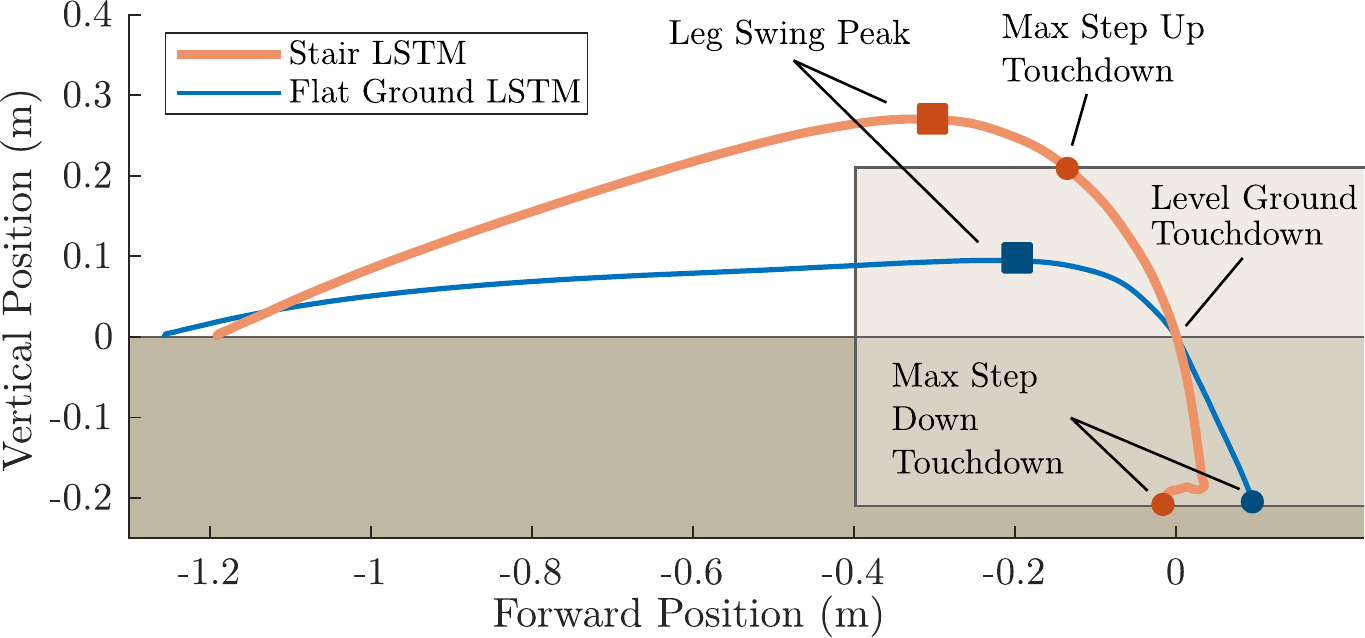}
\caption{Swing foot paths for the stair trained policy and the flat ground policy overlaid on example step ups and step downs.}
\label{fig:swing_leg_path}
\end{subfigure}

\begin{subfigure}[b]{1.0\columnwidth}
\centering
\vspace{5pt} 
\includegraphics[width=0.98\columnwidth]{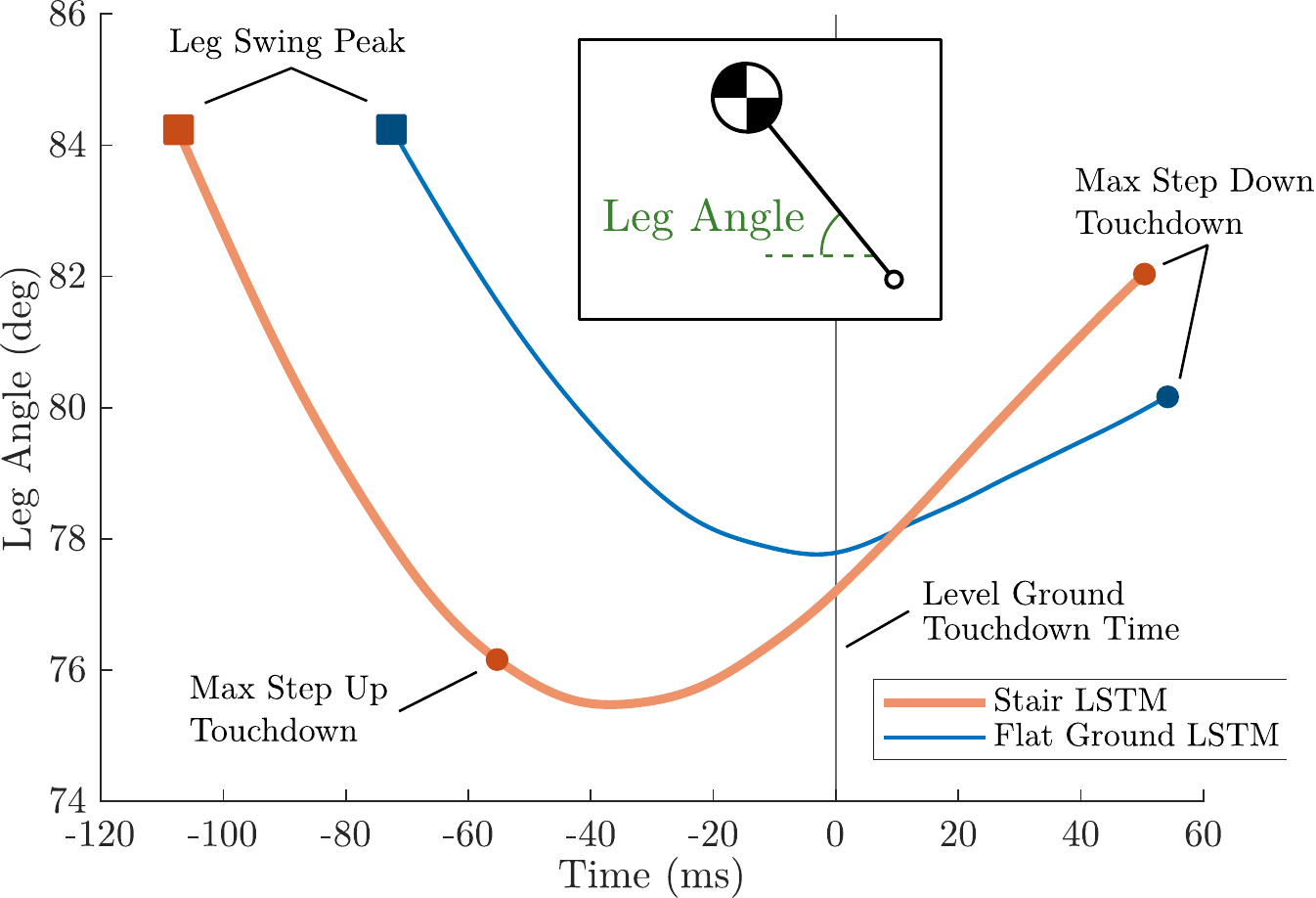}
\caption{The leg angle between the robot body and the swing foot as the foot descends toward touchdown.}
\label{fig:swing_leg_angle}
\end{subfigure}
\label{fig:swing_leg}
\caption{A comparison of the swing foot motion of the Stair LSTM policy and the Flat Ground LSTM policy while locomoting at 1.0 m/s. There is a significant change in the leg swing policy as a result of training on randomized stairs. The most significant changes are higher foot clearance, a steeper foot descent and a faster leg angle retraction rate.}
\end{figure}

To understand the change the stair terrain makes in the foot swing path we compare the result of a Flat Ground LSTM policy and a Stair LSTM policy when they encounter a drop step.
The foot swing path during a drop step lets us see where the policy would place the foot if it had encountered a step up or a step down.
Fig. \ref{fig:swing_leg_path} shows the foot swing path of these two policies relative to the ground. 
We can see that the Stair LSTM policy takes a much higher step compared to the Flat Ground LSTM policy which gives it additional clearance so it can step up onto a large step.
A second interesting observation is the steeper path of the swing foot for the Stair LSTM policy.
The swing foot only moves forward 14 cm while it is in the height range where it may encounter the front face of a step up.
We hypothesize this is a strategy that prevents the foot from stubbing the toe too hard on the front face of a stair and causing the robot to trip forward.

A second viewpoint to understand leg swing motion is to look at the leg swing retraction.
In humans and in bipedal birds it is observed that the swing leg is swung backwards, relative to the body, towards the ground near the end of stance \cite{poggensee2014characterizing, Daley2006RoughTerrain}.
This has the benefits of reducing the velocity of the foot relative to the ground and thus reducing the impact \cite{blum2010swing} as well as improving ground height disturbance rejection by automatically varying the leg touchdown angle \cite{seyfarth2003swing}.

Our training procedure does not explicitly incentivize the policy to exhibit these leg swing retraction behaviors, but we do see them emerge as shown in Fig. \ref{fig:swing_leg_angle}.
This figure shows the angle of the swing foot relative to the body between the peak of leg swing and contact with the maximum step down.
The Stair LSTM policy has a faster leg retraction rate compared to the Flat Ground LSTM policy.
With only this data we cannot say if this retraction profile is optimal or even if it is the cause of the improved performance on stairs.
However, the fact that there is a significant change in the leg retraction profile as a result of training on stairs is an interesting observation.

\subsubsection{Ground Reaction Forces}

\begin{figure*}
\centering
\includegraphics[width=0.75\textwidth]{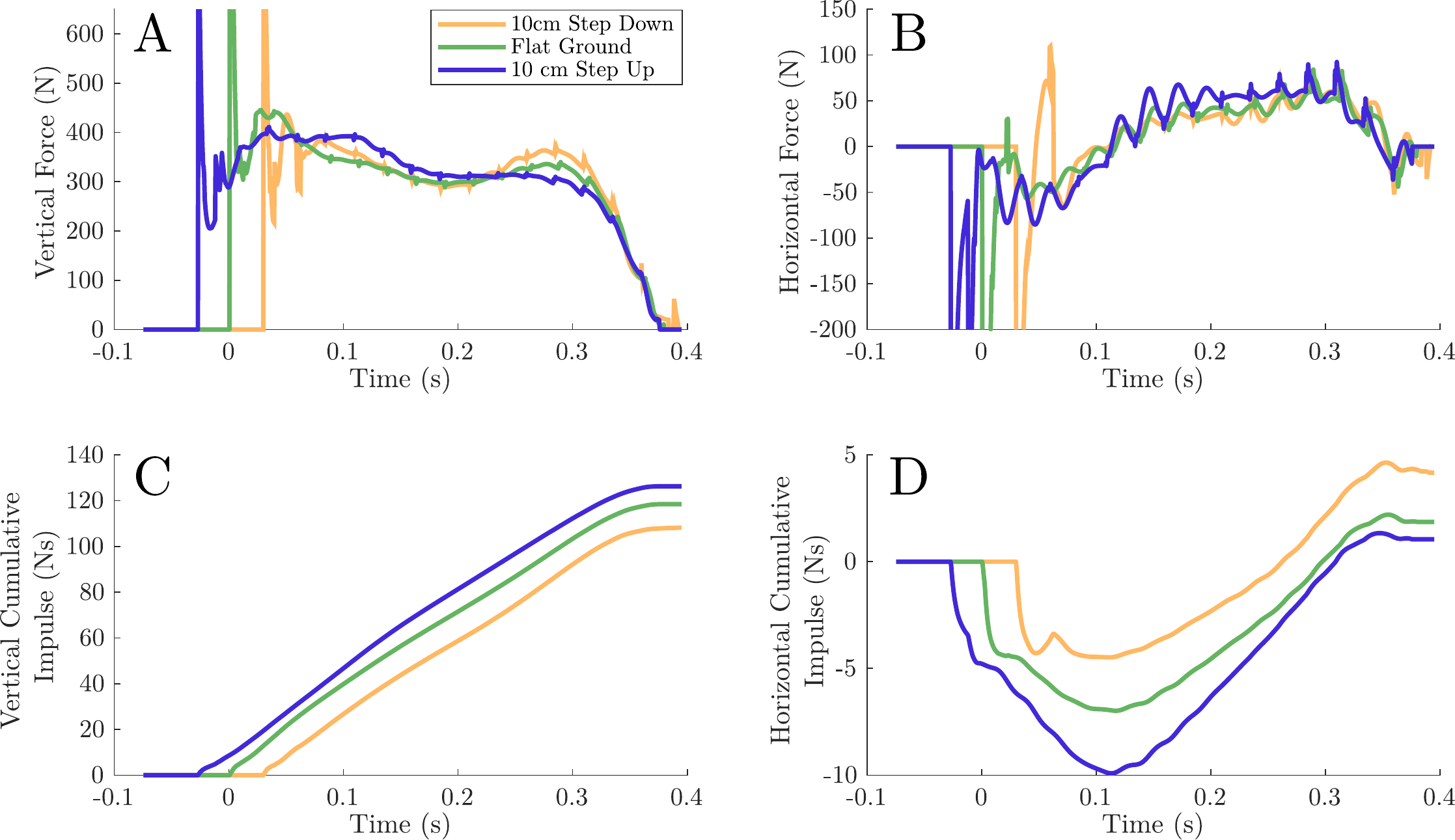}
\caption{The ground reaction forces and cumulative impulses of a Stair LSTM policy when it encounters varying ground height. The peak vertical force (A) after the impact remain roughly equivalent while the force in the second half of stance is modulated. The horizontal force (B) shows oscillations that match the frequency of the learned policy execution rate. This may be the policy controlling the body's attitude. The total vertical impulse (C) shows the expected result of a larger impulse stepping up and a smaller one stepping down. The horizontal impulse (D) shows a result that is predicted by leg swing retraction. When stepping down the foot is shifted backwards relative to the body which results in net acceleration forward which is shown here by a positive horizontal impulse.}
\label{fig:grf_plot}
\end{figure*}

Once the robot's foot has touched down its control authority is limited due to the underactuated nature of bipedal locomotion.
However, the robot still has a significant amount of control through the ground reaction force.
To understand how the Stair LSTM policy reacts to a 10 cm step up or down we plot the horizontal and vertical ground reaction forces in the sagittal plane in Fig. \ref{fig:grf_plot}.
At the beginning of stance there is a large spike in force that dwarfs the normal forces during stance.
The force value during this spike is largely defined by the tuning of the simulation contact model so it is not of primary interest to understanding the behavior of the policy.
The first interesting thing we see in subplot A is that the maximum nominal leg force is held relatively constant which is a predicted result of a well adjusted leg swing policy \cite{Birn-Jeffery3786}.
Second we see that the magnitude of the second hump in the double humped ground reaction forces is increased in the step down and decreased in the step up.
In the horizontal forces (subplot B) we see an oscillating signal where the oscillations match the frequency of policy evaluation.
We hypothesize that this is the policy working to control the attitude of the pelvis.
Prioritizing body attitude over forward velocity would be similar to the explicit priorities during single stance in Virtual Model Control \cite{Pratt2001_VMC}.
The lower two subplots (C and D) show the cumulative impulse in the vertical and forward directions throughout the stance phase.
We can see that the step up applies a larger vertical impulse and the step down a smaller vertical impulse.
This agrees with the intuition that the robot should apply a smaller vertical impulse to lower itself down a step compared to lifting itself up a step.
The horizontal impulse tells us if the robot speeds up or slows down in the forward direction during the stance phase.
We see that the step down results in a significantly larger forward impulse and the step up reduces the vertical impulse very slightly.
This aligns with the predicted behavior from a well tuned swing leg retraction policy.

\subsection{Hardware}

The recurrent policies transferred to hardware without any notable difficulties. We were able to take the robot for a walk around a large university campus using a randomly selected Stair LSTM policy and attempt to climb the staircases we came across. We observed robust and error-correcting behavior, as well as successful and repeatable stair ascents and descents. In addition, we noted robustness to uneven terrain, logs, and curbs, none of which were modeled in training. The policy was similarly robust to inclines and deformable terrain, demonstrated by a walk through a wet grass field and up a small hill. These experiments can be seen in our submission video \footnote{[Web link to submission video: \href{https://youtu.be/MPhEmC6b6XU}{youtu.be/MPhEmC6b6XU}]}, and a still image of one such experiment can be seen in Fig. \ref{fig:real_stairs}.

In addition to testing one-off terrains all over the university campus, we ran ten trials ascending stairs, and ten trials descending stairs on an outdoor real-world staircase. We recorded an 80\% success rate in ascending stairs using the selected Stair LSTM policy, and a 100\% success rate in descending stairs. A full video of this trial can be seen in our attachment to this submission. We note that the learned behavior is robust to missteps, and can quickly recover from mistakes, though the policy is not completely infallible and will fall if it makes a particularly egregious error. This experiment can be seen in our supplemental video \footnote{[Web link to supplemental video: \href{https://youtu.be/nuhHiKEtaZQ}{youtu.be/nuhHiKEtaZQ}]}.
The blind, proprioceptive learned strategy appears to rely on a solid stair face; evaluating policies on slatted stairs in simulation resulted in a much higher failure rate, pointing to the limits of such an approach. Even when explicitly included in training, slatted stairs tended to trip up policies on ascent. By contrast, stairs with randomly inclined steps (e.g., ones where each step had a unique pitch and roll orientation) did not seem to be difficult for ascent or descent. Likewise, approaching and ascending stairs at an angle did not seem to be an issue for policies.

\section{Conclusion} 
\label{sec:conclusion}

In this work, we have motivated the desirability of a highly robust but blind walking controller, and demonstrated that such a blind bipedal walking controller is capable of climbing a wide variety of real-world stairs. In addition, we note that producing such a controller requires very little modification to an existing training pipeline \cite{siekmann2020simtoreal}, and in particular no stair-specific reward terms; simply adding stairs to the environment with no further information is sufficient for learning stair-capable control policies. An important requirement of this learned ability appears to be a memory mechanism of some kind, probably due to the partially observable nature of the task of walking through unknown terrain while blind. 
In future work, it will be interesting to investigate how vision can be most effectively used to improve the efficiency and/or performance of a blind bipedal robot. Further, this work has demonstrated surprising capabilities for blind locomotion and leaves open the question of where the limits lie. 


 \section*{Acknowledgments}
 This work was partially supported by NSF grant IIS-1849343 and DARPA contract W911NF-16-1-0002. Special thanks to Intel for access to the vLab cluster, and to Yesh Godse, Jeremy Dao, and Helei Duan for advice and guidance.

\section*{Appendix}

\subsection*{Reward Function}

\begin{figure}[!h]
\centering
\includegraphics[width=\columnwidth]{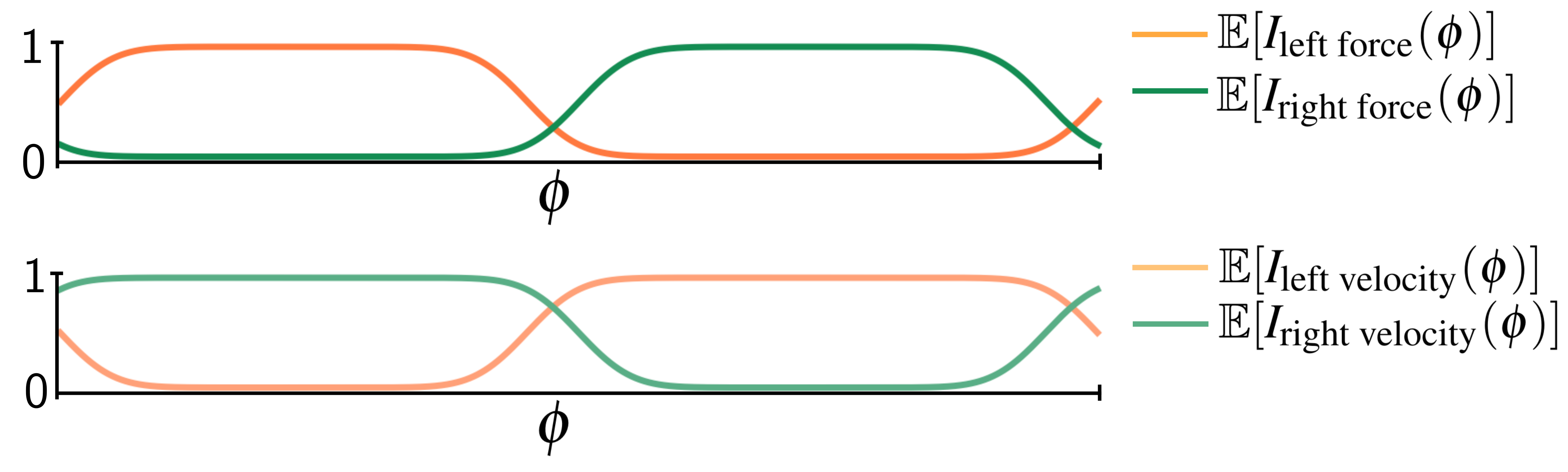}
\caption{By alternatingly punishing foot forces during a `stance' phase to teach the policy to lift the foot, and punishing foot velocities during a `swing' phase to teach the policy to place the foot on the ground, we can construct a foundation on which to learn simple walking behavior. Following in the path of previous work, we define these cyclic coefficents as random indicator functions of the phase, and take their expectation.}
\label{fig:reward_coefficent}
\end{figure}

To briefly review the approach taken in \cite{siekmann2020simtoreal}, we wish to take advantage of the complementary nature of foot forces and foot velocities during locomotion to construct a reward function which will punish one and allow the other, and vice versa, at key intervals during the gait. We use a probabilistic framework to represent uncertainty around the timings of these intervals. More specifically, we make use of a binary-valued random indicator function $I_i(\phi)$ for each quantity $q_i$ which we wish to penalize at some time during the gait cycle. This indicator function is likely to be 1 during the interval in which it is active, and likely to be 0 during intervals in which it is not active. The distribution of this binary-valued random function is defined via the Von Mises distribution; for a more comprehensive description, see \cite{siekmann2020learning}. 
In addition, rather than use the actual random variable in the reward we instead opt to use its expectation for more stable learning; see Fig. \ref{fig:reward_coefficent} for a plot of this expectation. 

Our full reward function is as follows;

\begin{equation}
R(s,\phi) = 1 - \mathbb{E}[\rho(s,\phi)]
\end{equation}

Which is to say, our reward is the difference of a bias and the expectation of a probabilistic penalty term $\rho(s, \phi)$ as described in \cite{siekmann2020simtoreal}. See Table \ref{table:rewardspec} for detailed information on the exact quantities and weightings used.

\bgroup
\def\arraystretch{1.2}
\begin{table}[!h]
\centering
\begin{tabular}{|l|l|}
\hline
Weight & Cost Component \\ \hline
0.140 &  $1 - \mathbb{E}[I_{\text{left force}}(\phi)] \cdot \exp(-.01\|F_l\|)$ \\ \hline
0.140 &  $1 - \mathbb{E}[I_{\text{right force}}(\phi)] \cdot \exp(-.01\|F_r\|)$ \\ \hline
0.140 &  $1 - \mathbb{E}[I_{\text{left velocity}}(\phi)] \cdot \exp(-\|v_l\|)$ \\ \hline
0.140 &  $1 - \mathbb{E}[I_{\text{right velocity}}(\phi)] \cdot \exp(-\|v_r\|)$ \\ \hline
0.140 & $1-\exp(-\epsilon_o)$ \\ \hline
0.140 & $1-\exp(-| \dot{x}_\text{desired} - \dot{x}_\text{actual}|)$ \\ \hline
0.078 & $1-\exp(-| \dot{y}_\text{desired} - \dot{y}_\text{actual}|)$ \\ \hline
0.028 & $1-\exp(-5\cdot\|a_t - a_{t-1}\|) $ \\ \hline  
0.028 & $1-\exp(-0.05 \cdot \|\tau\|) $ \\ \hline
0.028 & $1- \exp(-0.1 (\|\text{pelvis}_\text{rot}\| + \|\text{pelvis}_\text{acc}\|)) $ \\ \hline
\end{tabular}
\caption{The cost terms which are summed together to compose the expected penalty, $\mathbb{E}[\rho(s, \phi)]$. Terms involving an expectation of a variable $I_i(\phi)$ vary over the course of the gait cycle, with the goal of penalizing foot forces and foot velocities at key intervals to teach the policy to lift and place the feet periodically in order to walk. Other terms exist for the sake of commanding the policy to move forward, backwards, or sideways, or turn the robot to face a desired heading. Finally, the remaining terms exist to reduce behaviors which are shaky and thus unlikely to work well on hardware.}
\label{table:rewardspec}
\end{table}
\egroup


We define $F_l$ and $F_r$ as the vectors of translational forces applied to the left and right foot, and $v_l$ and $v_r$ similarly as the vectors of left and right foot velocities. To maintain a steady orientation, an orientation error $\epsilon_o$ is used, which is equal to,
\begin{equation}
\epsilon_o = 3(1 - \hat{\mathpzc{q}}^T\mathpzc{q}_{\:\text{body}})^2 + 10\left((1 - \hat{\mathpzc{q}}^T\mathpzc{q}_{\:\text{l}})^2 + (1 - \hat{\mathpzc{q}}^T\mathpzc{q}_{\:\text{r}})^2\right)
\end{equation}
where $\mathpzc{q}_{\:\text{l}}$ and $\mathpzc{q}_{\:\text{r}}$ are the quaternion orientations of the left and right foot, $\mathpzc{q}_{\:\text{body}}$ is the quaternion orientation of the pelvis, and $\hat{\mathpzc{q}}$ is a desired orientation (for our purposes, fixed to be always be facing straight ahead).

The quantities $\dot{x}_\text{desired}$ and $\dot{y}_\text{desired}$ correspond to a commanded translational speed, while $\dot{x}_\text{actual}$ and $\dot{y}_\text{actual}$ are the actual translational speed of the robot. The term $\text{pelvis}_\text{rot}$ represents the angular velocity while $\text{pelvis}_\text{acc}$ represents translational acceleration; these terms are used in the cost component to reduce the shakiness of locomotion behavior. The terms $a_t$ and $a_{t-1}$ refer to the current timestep's action and the previous timestep's action, and their use in the cost component is to encourage smooth behaviors. The term $\tau$ is the vector of net torques applied to the joints, and its use in the cost component is intended to encourage energy efficient gaits.

\bibliographystyle{unsrtnat}
\bibliography{references}

\begin{thebibliography}{29}
\providecommand{\natexlab}[1]{#1}
\providecommand{\url}[1]{\texttt{#1}}
\expandafter\ifx\csname urlstyle\endcsname\relax
  \providecommand{\doi}[1]{doi: #1}\else
  \providecommand{\doi}{doi: \begingroup \urlstyle{rm}\Url}\fi

\bibitem[Lee et~al.(2020)Lee, Hwangbo, Wellhausen, Koltun, and
  Hutter]{lee2020learning}
Joonho Lee, Jemin Hwangbo, Lorenz Wellhausen, Vladlen Koltun, and Marco Hutter.
\newblock \href{http://dx.doi.org/10.1126/scirobotics.abc5986}{Learning
  quadrupedal locomotion over challenging terrain}.
\newblock \emph{Science Robotics}, 5\penalty0 (47):\penalty0 eabc5986, Oct
  2020.
\newblock ISSN 2470-9476.
\newblock \doi{10.1126/scirobotics.abc5986}.

\bibitem[Bledt et~al.(2018)Bledt, Powell, Katz, Di~Carlo, Wensing, and
  Kim]{bledt2018cheetah}
Gerardo Bledt, Matthew~J Powell, Benjamin Katz, Jared Di~Carlo, Patrick~M
  Wensing, and Sangbae Kim.
\newblock \href{https://ieeexplore.ieee.org/abstract/document/8593885}{MIT
  Cheetah 3: Design and control of a robust, dynamic quadruped robot}.
\newblock In \emph{2018 IEEE/RSJ International Conference on Intelligent Robots
  and Systems (IROS)}, pages 2245--2252. IEEE, 2018.

\bibitem[Moore and Buehler(2001)]{moore2001stable}
EZ~Moore and M~Buehler.
\newblock \href{https://apps.dtic.mil/sti/citations/ADA438777}{Stable stair
  climbing in a simple hexapod robot}.
\newblock Technical report, McGill Research Centre for Intelligent Machines,
  2001.

\bibitem[Park et~al.(2012)Park, Ramezani, and Grizzle]{park2012finite}
Hae-Won Park, Alireza Ramezani, and Jessy~W Grizzle.
\newblock \href{https://ieeexplore.ieee.org/abstract/document/6399609}{A
  finite-state machine for accommodating unexpected large ground-height
  variations in bipedal robot walking}.
\newblock \emph{IEEE Transactions on Robotics}, 29\penalty0 (2):\penalty0
  331--345, 2012.

\bibitem[Gutmann et~al.(2004)Gutmann, Fukuchi, and Fujita]{gutmann2004stair}
J-S Gutmann, Masaki Fukuchi, and Masahiro Fujita.
\newblock \href{https://ieeexplore.ieee.org/abstract/document/1389593}{Stair
  climbing for humanoid robots using stereo vision}.
\newblock In \emph{2004 IEEE/RSJ International Conference on Intelligent Robots
  and Systems (IROS)(IEEE Cat. No. 04CH37566)}, volume~2, pages 1407--1413.
  IEEE, 2004.

\bibitem[Albert et~al.(2001)Albert, Suppa, and Gerth]{albert2001detection}
Amos Albert, Michael Suppa, and Wilfried Gerth.
\newblock \href{https://ieeexplore.ieee.org/abstract/document/936909}{Detection
  of stair dimensions for the path planning of a bipedal robot}.
\newblock In \emph{2001 IEEE/ASME International Conference on Advanced
  Intelligent Mechatronics. Proceedings (Cat. No. 01TH8556)}, volume~2, pages
  1291--1296. IEEE, 2001.

\bibitem[Michel et~al.(2007)Michel, Chestnutt, Kagami, Nishiwaki, Kuffner, and
  Kanade]{michel2007gpu}
Philipp Michel, Joel Chestnutt, Satoshi Kagami, Koichi Nishiwaki, James
  Kuffner, and Takeo Kanade.
\newblock
  \href{https://ieeexplore.ieee.org/abstract/document/4399104}{GPU-accelerated
  real-time 3D tracking for humanoid locomotion and stair climbing}.
\newblock In \emph{2007 IEEE/RSJ International Conference on Intelligent Robots
  and Systems}, pages 463--469. IEEE, 2007.

\bibitem[Caron et~al.(2019)Caron, Kheddar, and Tempier]{caron2019stair}
St{\'e}phane Caron, Abderrahmane Kheddar, and Olivier Tempier.
\newblock \href{https://ieeexplore.ieee.org/abstract/document/8794348}{Stair
  climbing stabilization of the HRP-4 humanoid robot using whole-body
  admittance control}.
\newblock In \emph{2019 International Conference on Robotics and Automation
  (ICRA)}, pages 277--283. IEEE, 2019.

\bibitem[Robotics()]{agility2019video}
Agility Robotics.
\newblock \href{https://youtu.be/qV-92Bq96Co}{Cassie: Dynamic Planning on
  Stairs}.

\bibitem[Figliolini and Ceccarelli(2001)]{figliolini2001climbing}
Giorgio Figliolini and Marco Ceccarelli.
\newblock \href{https://ieeexplore.ieee.org/abstract/document/933261}{Climbing
  stairs with EP-WAR2 biped robot}.
\newblock In \emph{Proceedings 2001 ICRA. IEEE International Conference on
  Robotics and Automation (Cat. No. 01CH37164)}, volume~4, pages 4116--4121.
  IEEE, 2001.

\bibitem[Focchi et~al.(2020)Focchi, Orsolino, Camurri, Barasuol, Mastalli,
  Caldwell, and Semini]{Focchi2020variablesensor}
Michele Focchi, Romeo Orsolino, Marco Camurri, Victor Barasuol, Carlos
  Mastalli, Darwin~G. Caldwell, and Claudio Semini.
\newblock \emph{\href{https://doi.org/10.1007/978-3-030-22327-4_9}{Heuristic
  Planning for Rough Terrain Locomotion in Presence of External Disturbances
  and Variable Perception Quality}}, volume 132, pages 165--209.
\newblock Springer International Publishing, Cham, 2020.
\newblock ISBN 978-3-030-22327-4.
\newblock \doi{10.1007/978-3-030-22327-4_9}.

\bibitem[Xie et~al.(2018)Xie, Berseth, Clary, Hurst, and van~de
  Panne]{xie2018feedback}
Zhaoming Xie, Glen Berseth, Patrick Clary, Jonathan Hurst, and Michiel van~de
  Panne.
\newblock \href{https://ieeexplore.ieee.org/abstract/document/8593722}{Feedback
  control for cassie with deep reinforcement learning}.
\newblock In \emph{2018 IEEE/RSJ International Conference on Intelligent Robots
  and Systems (IROS)}, pages 1241--1246. IEEE, 2018.

\bibitem[Siekmann et~al.(2021)Siekmann, Godse, Fern, and
  Hurst]{siekmann2020simtoreal}
Jonah Siekmann, Yesh Godse, Alan Fern, and Jonathan Hurst.
\newblock \href{https://arxiv.org/abs/2011.01387}{Sim-to-Real Learning of All
  Common Bipedal Gaits via Periodic Reward Composition}.
\newblock In \emph{IEEE International Conference on Robotics and Automation
  (ICRA)}, 2021.

\bibitem[Sutton and Barto(2018)]{Sutton2018}
Richard~S Sutton and Andrew~G Barto.
\newblock
  \emph{\href{http://incompleteideas.net/book/the-book.html}{Reinforcement
  learning: An introduction}}.
\newblock MIT press, 2018.

\bibitem[Peng and van~de Panne(2016)]{peng2016learning}
Xue~Bin Peng and Michiel van~de Panne.
\newblock \href{https://dl.acm.org/doi/abs/10.1145/3099564.3099567}{Learning
  Locomotion Skills Using DeepRL: Does the Choice of Action Space Matter?}
\newblock \emph{CoRR}, abs/1611.01055, 2016.

\bibitem[{Peng} et~al.(2018){Peng}, {Andrychowicz}, {Zaremba}, and
  {Abbeel}]{peng2019simtoreal}
X.~B. {Peng}, M.~{Andrychowicz}, W.~{Zaremba}, and P.~{Abbeel}.
\newblock
  \href{https://ieeexplore.ieee.org/abstract/document/8460528}{Sim-to-Real
  Transfer of Robotic Control with Dynamics Randomization}.
\newblock In \emph{2018 IEEE International Conference on Robotics and
  Automation (ICRA)}, pages 3803--3810, 2018.
\newblock \doi{10.1109/ICRA.2018.8460528}.

\bibitem[Hochreiter and Schmidhuber(1997)]{hochreiter1997long}
Sepp Hochreiter and J{\"u}rgen Schmidhuber.
\newblock
  \href{https://www.mitpressjournals.org/doi/abs/10.1162/neco.1997.9.8.1735}{Long
  short-term memory}.
\newblock \emph{Neural computation}, 9\penalty0 (8):\penalty0 1735--1780, 1997.

\bibitem[Heess et~al.(2015)Heess, Hunt, Lillicrap, and
  Silver]{heess2015memorybased}
Nicolas Heess, Jonathan~J Hunt, Timothy~P Lillicrap, and David Silver.
\newblock \href{https://arxiv.org/abs/1512.04455}{Memory-based control with
  recurrent neural networks}, 2015.

\bibitem[Siekmann et~al.(2020)Siekmann, Valluri, Dao, Bermillo, Duan, Fern, and
  Hurst]{siekmann2020learning}
Jonah Siekmann, Srikar Valluri, Jeremy Dao, Lorenzo Bermillo, Helei Duan, Alan
  Fern, and Jonathan Hurst.
\newblock
  \href{https://roboticsconference.org/2020/program/papers/31.html}{Learning
  memory-based control for human-scale bipedal locomotion}.
\newblock In \emph{Proceedings of Robotics: Science and Systems}, July 2020.

\bibitem[Schulman et~al.(2017)Schulman, Wolski, Dhariwal, Radford, and
  Klimov]{schulman2017proximal}
John Schulman, Filip Wolski, Prafulla Dhariwal, Alec Radford, and Oleg Klimov.
\newblock {\href{https://arxiv.org/abs/1707.06347}{Proximal Policy Optimization
  Algorithms}}, 2017.

\bibitem[Adbolhosseini et~al.(2019)Adbolhosseini, Ling, Xie, Peng, and van~de
  Panne]{mig2019symmetry}
Farzad Adbolhosseini, Hung~Yu Ling, Zhaoming Xie, Xue~Bin Peng, and Michiel
  van~de Panne.
\newblock \href{https://dl.acm.org/doi/10.1145/3359566.3360070}{On Learning
  Symmetric Locomotion}.
\newblock In \emph{Proc. ACM SIGGRAPH Motion, Interaction, and Games (MIG
  2019)}, 2019.

\bibitem[Todorov et~al.(2012)Todorov, Erez, and Tassa]{todorov2012mujoco}
Emanuel Todorov, Tom Erez, and Yuval Tassa.
\newblock \href{https://ieeexplore.ieee.org/abstract/document/6386109}{Mujoco:
  A physics engine for model-based control}.
\newblock In \emph{2012 IEEE/RSJ International Conference on Intelligent Robots
  and Systems}, pages 5026--5033. IEEE, 2012.

\bibitem[Kingma and Ba(2014)]{kingma2014adam}
Diederik~P Kingma and Jimmy Ba.
\newblock \href{https://arxiv.org/abs/1412.6980}{Adam: A method for stochastic
  optimization}.
\newblock \emph{arXiv preprint arXiv:1412.6980}, 2014.

\bibitem[Poggensee et~al.(2014)Poggensee, Sharbafi, and
  Seyfarth]{poggensee2014characterizing}
KL~Poggensee, MA~Sharbafi, and A~Seyfarth.
\newblock \href{https://doi.org/10.1142/9789814623353_0044}{Characterizing
  swing-leg retraction in human locomotion}.
\newblock In \emph{Mobile Service Robotics}, pages 377--384. World Scientific,
  2014.

\bibitem[Daley and Biewener(2006)]{Daley2006RoughTerrain}
Monica~A. Daley and Andrew~A. Biewener.
\newblock \href{https://www.pnas.org/content/103/42/15681}{Running over rough
  terrain reveals limb control for intrinsic stability}.
\newblock \emph{Proceedings of the National Academy of Sciences}, 103\penalty0
  (42):\penalty0 15681--15686, 2006.
\newblock ISSN 0027-8424.
\newblock \doi{10.1073/pnas.0601473103}.

\bibitem[Blum et~al.(2010)Blum, Lipfert, Rummel, and Seyfarth]{blum2010swing}
Yvonne Blum, Susanne~W Lipfert, Juergen Rummel, and Andr{\'e} Seyfarth.
\newblock
  \href{https://iopscience.iop.org/article/10.1088/1748-3182/5/2/026006/meta}{Swing
  leg control in human running}.
\newblock \emph{Bioinspiration \& biomimetics}, 5\penalty0 (2):\penalty0
  026006, 2010.

\bibitem[Seyfarth et~al.(2003)Seyfarth, Geyer, and Herr]{seyfarth2003swing}
Andr{\'e} Seyfarth, Hartmut Geyer, and Hugh Herr.
\newblock \href{https://jeb.biologists.org/content/206/15/2547.short}{Swing-leg
  retraction: a simple control model for stable running}.
\newblock \emph{Journal of Experimental Biology}, 206\penalty0 (15):\penalty0
  2547--2555, 2003.

\bibitem[Birn-Jeffery et~al.(2014)Birn-Jeffery, Hubicki, Blum, Renjewski,
  Hurst, and Daley]{Birn-Jeffery3786}
Aleksandra~V. Birn-Jeffery, Christian~M. Hubicki, Yvonne Blum, Daniel
  Renjewski, Jonathan~W. Hurst, and Monica~A. Daley.
\newblock
  \href{https://jeb.biologists.org/content/217/21/3786}{Don{\textquoteright}t
  break a leg: running birds from quail to ostrich prioritise leg safety and
  economy on uneven terrain}.
\newblock \emph{Journal of Experimental Biology}, 217\penalty0 (21):\penalty0
  3786--3796, 2014.
\newblock ISSN 0022-0949.
\newblock \doi{10.1242/jeb.102640}.

\bibitem[\href{https://doi.org/10.1177/02783640122067309}{Jerry Pratt and
  Chee-Meng Chew and Ann Torres and Peter Dilworth and Gill
  Pratt}(2001)]{Pratt2001_VMC}
\href{https://doi.org/10.1177/02783640122067309}{Jerry Pratt and Chee-Meng Chew
  and Ann Torres and Peter Dilworth and Gill Pratt}.
\newblock Virtual model control: An intuitive approach for bipedal locomotion.
\newblock \emph{The International Journal of Robotics Research}, 20\penalty0
  (2):\penalty0 129--143, 2001.
\newblock \doi{10.1177/02783640122067309}.

\end{thebibliography}

\end{document}